# Generalized Fisher Score for Feature Selection

**Quanquan Gu**
Dept. of Computer Science
University of Illinois at
Urbana-Champaign
Urbana, IL 61801, US
qgu3@illinois.edu

**Zhenhui Li**
Dept. of Computer Science
University of Illinois at
Urbana-Champaign
Urbana, IL 61801, US
zli28@uiuc.edu

**Jiawei Han**
Dept. of Computer Science
University of Illinois at
Urbana-Champaign
Urbana, IL 61801, US
hanj@cs.uiuc.edu

## Abstract

Fisher score is one of the most widely used supervised feature selection methods. However, it selects each feature independently according to their scores under the Fisher criterion, which leads to a suboptimal subset of features. In this paper, we present a generalized Fisher score to jointly select features. It aims at finding an subset of features, which maximize the lower bound of traditional Fisher score. The resulting feature selection problem is a mixed integer programming, which can be reformulated as a quadratically constrained linear programming (QCLP). It is solved by cutting plane algorithm, in each iteration of which a multiple kernel learning problem is solved alternatively by multivariate ridge regression and projected gradient descent. Experiments on benchmark data sets indicate that the proposed method outperforms Fisher score as well as many other state-of-the-art feature selection methods.

## 1 Introduction

High-dimensional data in the input space is usually not good for classification due to the *curse of dimensionality* [15]. It significantly increases the time and space complexity for processing the data. Moreover, in the presence of many irrelevant and/or redundant features, learning methods tend to over-fit and become less interpretable. A common way to resolve this problem is feature selection [21, 5], which reduces the dimensionality by selecting a subset of features from the input feature set. It is often used to reduce the computational cost and remove irrelevant and redundant features for problems with high dimensional data.

Generally speaking, feature selection methods can be categorized into three families: filter-based, wrapper-based and embedded methods [5]. Filter-based methods rank the features as a pre-processing step prior to the learning algorithm, and select those features with high ranking scores. Wrapper-based methods score the features using the learning algorithm that will ultimately be employed. Embedded methods combine feature selection with the learning algorithm. The design of embedded method is tightly coupled with a specific learning algorithm, which in turn limits its application to other learning algorithms. In our study, we focus on filter-based methods for supervised feature selection.

Filter-based feature selection is usually cast into a binary selection of features which maximizes some performance criterion. In the past decade, a number of performance criteria have been proposed for filter-based feature selection, such as mutual information [10], Fisher score [15], ReliefF [17], Laplacian score [6], Hilbert Schmidt Independence Criterion (HSIC) [18] and Trace Ratio criterion [12], among which Fisher score is one of the most widely used criteria for supervised feature selection due to its general good performance. In detail, given a set of $d$ features, denoted by $\mathcal{S}$, the goal of filter based feature selection is to choose a subset of $m < d$ features, denoted by $\mathcal{T}$, which maximizes some criterion $F$,

$$\mathcal{T}^* = \arg\max_{\mathcal{T} \subseteq \mathcal{S}} F(\mathcal{T}), \text{s.t.} |\mathcal{T}| = m \qquad (1)$$

where $|\cdot|$ is the cardinality of a set. Eq.(1) is a combinatorial optimization problem. Finding the global optimal solution for Eq.(1) is NP hard. One common heuristic approach addresses this issue by first computing a score for each feature independently according to the criterion $F$, and then selecting the top-$m$ ranked features with high scores. However, the features selected by the heuristic algorithm is suboptimal. On the one hand, since the heuristic algorithm computes the score of each feature individually, it neglects the combination of features, which means evaluating two or more than two features together. For example, it could be the case that the scores of feature $a$ and fea-ture $b$ are both low, but the score of the combination $ab$ is very high. In this case, the heuristic algorithm will discard feature $a$ and $b$, although they should be selected. On the other hand, they cannot handle redundant features. For instance, the scores of feature $a$ and feature $b$ are very high, but they are highly correlated. In this case, the heuristic algorithm will select both feature $a$ and $b$, while either feature $a$ or $b$ could be eliminated without loss in the subsequent learning performance. In fact, many studies have shown that removing redundant features can result in performance improvement [1, 23, 14].

In this paper, in order to overcome the above problems, we present a generalized Fisher score for feature selection. Rather than selecting each feature individually, the proposed method selects a subset of features simultaneously. It aims to find a subset of features, which maximize the lower bound of traditional Fisher score. It is able to consider the combination of features, and eliminate the redundant features. The resulting feature selection problem is a mixed integer programming, which is further reformulated as a quadratically constrained linear programming (QCLP) [3]. It can be solved by cutting plane algorithm [8], in each iteration of which a multiple kernel learning problem is solved by multivariate ridge regression and projected gradient descent [16] alternatively. Experiments on benchmark data sets indicate that the proposed method outperforms many state of the art feature selection methods.

The remainder of this paper is organized as follows. In Section 2, we briefly review Fisher score. We present the generalized Fisher score in Section 3. The experiments on benchmark data sets are demonstrated in Section 4. Finally, we draw a conclusion in Section 5.

## 1.1 Notation

The generic problem of supervised feature selection is as follows. Given a data set $\{(\mathbf{x}_i, y_i)\}_{i=1}^n$ where $\mathbf{x}_i \in \mathbb{R}^d$ and $y_i \in \{1, 2, \ldots, c\}$, we aim to find a feature subset of size $m$ which contains the most informative features. We use $\mathbf{X} = [\mathbf{x}_1, \mathbf{x}_2, \ldots, \mathbf{x}_n] \in \mathbb{R}^{d \times n}$ to represent the data matrix. $\mathbf{x}^j$ denotes the $j$th row of $\mathbf{X}$. **1** is a vector of all ones with an appropriate length. **0** is a vector of all zeros. **I** is an identity matrix with an appropriate size. Without loss of generality, we assume that $\mathbf{X}$ has been centered with zero mean, i.e., $\sum_{i=1}^n \mathbf{x}_i = \mathbf{0}$.

## 2 A Brief Review of Fisher Score

In this section, we briefly review Fisher score [15] for feature selection, and discuss its shortcomings.

The key idea of Fisher score is to find a subset of features, such that in the data space spanned by the selected features, the distances between data points in different classes are as large as possible, while the distances between data points in the same class are as small as possible. In particular, given the selected $m$ features, the input data matrix $\mathbf{X} \in \mathbb{R}^{d \times n}$ reduces to $\mathbf{Z} \in \mathbb{R}^{m \times n}$. Then the *Fisher Score* is computed as follows,

$$F(\mathbf{Z}) = \text{tr}\left\{(\tilde{\mathbf{S}}_b)(\tilde{\mathbf{S}}_t + \gamma \mathbf{I})^{-1}\right\}, \qquad (2)$$

where $\gamma$ is a positive regularization parameter, $\tilde{\mathbf{S}}_b$ is called between-class scatter matrix, and $\tilde{\mathbf{S}}_t$ is called total scatter matrix, which are defined as

$$\begin{aligned}
\tilde{\mathbf{S}}_b &= \sum_{k=1}^c n_k(\tilde{\boldsymbol{\mu}}_k - \tilde{\boldsymbol{\mu}})(\tilde{\boldsymbol{\mu}}_k - \tilde{\boldsymbol{\mu}})^T \\
\tilde{\mathbf{S}}_t &= \sum_{i=1}^n (\mathbf{z}_i - \tilde{\boldsymbol{\mu}})(\mathbf{z}_i - \tilde{\boldsymbol{\mu}})^T,
\end{aligned} \qquad (3)$$

where $\tilde{\boldsymbol{\mu}}_k$ and $n_k$ are the mean vector and size of the $k$-th class respectively in the reduced data space, i.e., $\mathbf{Z}$, $\tilde{\boldsymbol{\mu}} = \sum_{k=1}^c n_k \tilde{\boldsymbol{\mu}}_k$ is the overall mean vector of the reduced data. Since $\tilde{\mathbf{S}}_t$ is usually singular, we add a perturbation term $\gamma \mathbf{I}$ to make it positive semi-definite.

Since there are $\binom{d}{m}$ candidate $\mathbf{Z}$'s out of $\mathbf{X}$, the feature selection problem is a combinatorial optimization problem and very challenging. To alleviate the difficulty, the widely used heuristic strategy [15, 6] is to compute a score for each feature independently according to the criterion $F$. In other word, it only considers $\mathbf{x}^j \in \mathbb{R}^{1 \times n}$. In this case, there are only $\binom{d}{1} = d$ candidates. Specifically, let $\mu_k^j$ and $\sigma_k^j$ be the mean and standard deviation of $k$-th class, corresponding to the $j$-th feature. Let $\mu^j$ and $\sigma^j$ denote the mean and standard deviation of the whole data set corresponding to the $j$-th feature. Then the Fisher score of the $j$-th feature is computed below,

$$F(\mathbf{x}^j) = \frac{\sum_{k=1}^c n_k(\mu_k^j - \mu^j)^2}{(\sigma^j)^2}, \qquad (4)$$

where $(\sigma^j)^2 = \sum_{k=1}^c n_k(\sigma_k^j)^2$. After computing the Fisher score for each feature, it selects the top-$m$ ranked features with large scores. Because the score of each feature is computed independently, the features selected by the heuristic algorithm is suboptimal. More importantly, as we mentioned before, the heuristic algorithm fails to select those features which have relatively low individual scores but a very high score when they are combined together as a whole. In addition, it cannot handle redundant features. This motivates us to propose a generalized Fisher score which can resolve these problems.

# 3 The Proposed Method

In this section, we first present an equivalent formulation of Fisher score, based on which we present our method.

## 3.1 Equivalent Formulation of Fisher Score

We introduce an indicator variable $\mathbf{p}$, where $\mathbf{p} = (p_1, \ldots, p_d)^T$ and $p_i \in \{0,1\}, i = 1, \ldots, d$, to represent whether a feature is selected or not. In order to indicate that $m$ features are selected, we constrain $\mathbf{p}$ by $\mathbf{p}^T \mathbf{1} = m$. Then the *Fisher Score* in Eq.(2) can be equivalently formulated as follows,

$$
\begin{aligned}
F(\mathbf{p}) &= \mathrm{tr}\{(\mathrm{diag}(\mathbf{p})\mathbf{S}_b \mathrm{diag}(\mathbf{p})) \\
&\quad (\mathrm{diag}(\mathbf{p})(\mathbf{S}_t + \gamma \mathbf{I})\mathrm{diag}(\mathbf{p}))^{-1}\}, \\
\text{s.t.} &\quad \mathbf{p} \in \{0,1\}^d, \mathbf{p}^T \mathbf{1} = m,
\end{aligned}
\quad (5)
$$

where $\mathrm{diag}(\mathbf{p})$ is a diagonal matrix whose diagonal elements are $p_i$'s, $\mathbf{S}_b$ and $\mathbf{S}_t$ are the between-class scatter matrix and total scatter matrix respectively, which are defined as

$$
\begin{aligned}
\mathbf{S}_b &= \sum_{k=1}^{c} n_k (\boldsymbol{\mu}_k - \boldsymbol{\mu})(\boldsymbol{\mu}_k - \boldsymbol{\mu})^T \\
\mathbf{S}_t &= \sum_{i=1}^{n} (\mathbf{x}_i - \boldsymbol{\mu})(\mathbf{x}_i - \boldsymbol{\mu})^T
\end{aligned}
\quad (6)
$$

where $\boldsymbol{\mu}_k$ and $n_k$ are mean vector and size of $k$th class respectively in the input data space, i.e., $\mathbf{X}$, $\boldsymbol{\mu} = \sum_{k=1}^{c} n_k \boldsymbol{\mu}_k$ is the overall mean vector of the original data.

## 3.2 Generalized Fisher Score

However, the problem in Eq.(5) is still not easy to maximize due to its combinatorial nature. In this paper, we turn to maximize its lower bound as follows.

**Theorem 3.1.** *The optimal value of the objective function in Eq.(5) is lower bounded by the optimal value of the objective function in the following problem,*

$$
\begin{aligned}
F(\mathbf{W}, \mathbf{p}) &= tr\{(\mathbf{W}^T diag(\mathbf{p})\mathbf{S}_b diag(\mathbf{p})\mathbf{W}) \\
&\quad (\mathbf{W}^T diag(\mathbf{p})(\mathbf{S}_t + \gamma \mathbf{I}) diag(\mathbf{p})\mathbf{W})^{-1}\}, \\
\text{s.t.} &\quad \mathbf{p} \in \{0,1\}^d, \mathbf{p}^T \mathbf{1} = m.
\end{aligned}
\quad (7)
$$

where $\mathbf{W} \in \mathbb{R}^{d \times c}$.

*Proof.* The key is to prove that for any feasible $\mathbf{p}$, the objective function of Eq.(5) is lower bounded by the objective function of Eq.(7). The detailed proof will be included in the longer version of this paper. □

We call the feature selection criterion in Eq.(7) as *Generalized Fisher Score*. It is easy to show that given $\mathbf{p}$, Eq.(7) can be seen as *Regularized Discriminant Analysis* (RDA) [15] in the reduced feature space, i.e., $\mathrm{diag}(\mathbf{p})\mathbf{X}$, which is a *Rayleigh Quotient* problem [4] (also known as Ratio Trace problem), and can be solved by eigen-decomposition. However, when $\mathbf{p}$ is also a variable, the problem is difficult to solve. Recent study [22] established the relationship between RDA and multi-variate linear regression problem, which provides a regression-based solution for RDA. This motivates us to solve the problem in Eq.(7) in a similar manner. In the following, we present a theorem, which establishes the equivalence relationship between the problem in Eq.(7) and the problem in Eq.(8).

**Theorem 3.2.** *The optimal $\mathbf{p}$ that maximize the problem in Eq.(7) are the same as the optimal $\mathbf{p}$ that minimize the following problem*

$$
\begin{aligned}
\min_{\mathbf{p},\mathbf{W}} &\quad \frac{1}{2}\|\mathbf{X}^T diag(\mathbf{p})\mathbf{W} - \mathbf{H}\|_F^2 + \frac{\gamma}{2}\|\mathbf{W}\|_F^2, \\
\text{s.t.} &\quad \mathbf{p} \in \{0,1\}^d, \mathbf{p}^T \mathbf{1} = m.
\end{aligned}
\quad (8)
$$

*where* $\mathbf{H} = [\mathbf{h}_1, \ldots, \mathbf{h}_c] \in \mathbb{R}^{n \times c}$, *and* $\mathbf{h}_k$ *is a column vector whose $i$-th entry is given by*

$$
h_{ik} = \begin{cases} \sqrt{\frac{n}{n_k}} - \sqrt{\frac{n_k}{n}}, & \text{if } \mathbf{y}_i = k \\ -\sqrt{\frac{n_k}{n}}, & \text{otherwise.} \end{cases}
\quad (9)
$$

*Proof.* The sketch of the proof is: for any feasible $\mathbf{W}$, the optimal $\mathbf{p}$ that maximizes the problem in Eq.(7) is the same as the optimal $\mathbf{p}$ that minimizes the problem in Eq.(8). It is built upon Lemma 4.1 in [22]. The detailed proof will be included in the longer version of this paper. □

Note that the above theorem holds under the condition that $\mathbf{X}$ is centered with zero mean. It is interesting to note that, in general, the optimal $\mathbf{W}$ for the optimization problem in Eq.(7) is different from the optimal $\mathbf{W}$ for the problem in Eq.(8).

## 3.3 The Dual Problem

According to Theorem 3.2, we can solve the multivariate ridge regression like problem in Eq.(8) instead of the Ratio Trace problem in Eq.(7). Let

$$
\mathbf{U} = \mathbf{X}^T \mathrm{diag}(\mathbf{p})\mathbf{W} - \mathbf{H}. \quad (10)
$$

The optimization problem in Eq.(8) is equivalent to the following optimization problem,

$$
\begin{aligned}
\min_{\mathbf{p},\mathbf{W}} &\quad \frac{1}{2}\|\mathbf{U}\|_F^2 + \frac{\gamma}{2}\|\mathbf{W}\|_F^2 \\
\text{s.t.} &\quad \mathbf{U} = \mathbf{X}^T \mathrm{diag}(\mathbf{p})\mathbf{W} - \mathbf{H} \\
&\quad \mathbf{p} \in \{0,1\}^d, \mathbf{p}^T \mathbf{1} = m.
\end{aligned}
\quad (11)
$$

We consider the dual problem of Eq.(11). The Lagrangian function of Eq.(11) is as follows,

$$L = \frac{1}{2}||\mathbf{U}||_F^2 + \frac{\gamma}{2}||\mathbf{W}||_F^2 - \text{tr}(\mathbf{V}^T(\mathbf{X}^T\text{diag}(\mathbf{p})\mathbf{W} - \mathbf{H} - \mathbf{U})), \quad (12)$$

where $\mathbf{V}$ is a Lagrangian multiplier. Taking the derivative of $L$ with respect to $\mathbf{U}$ and $\mathbf{W}$ and setting them to zero, we obtain

$$\frac{\partial L}{\partial \mathbf{U}} = \mathbf{U} + \mathbf{V} = 0$$
$$\frac{\partial L}{\partial \mathbf{W}} = \gamma \mathbf{W} - \text{diag}(\mathbf{p})\mathbf{X}\mathbf{V} = 0. \quad (13)$$

It follows that

$$\mathbf{U} = -\mathbf{V}$$
$$\mathbf{W} = \frac{1}{\gamma}\text{diag}(\mathbf{p})\mathbf{X}\mathbf{V}. \quad (14)$$

Thus we obtain the following dual problem of Eq.(11)

$$\min_{\mathbf{p}} \max_{\mathbf{V}} \quad \text{tr}(\mathbf{V}^T\mathbf{H}) - \frac{1}{2}\text{tr}(\mathbf{V}^T(\frac{1}{\gamma}\mathbf{X}^T\text{diag}(\mathbf{p})\mathbf{X} + \mathbf{I})\mathbf{V})$$
$$\text{s.t.} \quad \mathbf{p} \in \{0,1\}^d, \mathbf{p}^T\mathbf{1} = m. \quad (15)$$

For notational simplicity, we denote the objective function in Eq.(15) as

$$f(\mathbf{V}, \mathbf{p}) = \text{tr}(\mathbf{V}^T\mathbf{H}) - \frac{1}{2}\text{tr}(\mathbf{V}^T(\frac{1}{\gamma}\sum_{j=1}^{d}p_j\mathbf{K}_j + \mathbf{I})\mathbf{V}), \quad (16)$$

where $\mathbf{K}_j = (\mathbf{x}^j)^T\mathbf{x}^j$ and

$$\mathcal{P} = \{\mathbf{p}|\mathbf{p} \in \{0,1\}^d, \mathbf{p}^T\mathbf{1} = m\}. \quad (17)$$

Then Eq.(15) is simplified as

$$\min_{\mathbf{p}\in\mathcal{P}} \max_{\mathbf{V}} f(\mathbf{V}, \mathbf{p}). \quad (18)$$

By interchanging the order of $\min_{\mathbf{p}\in\mathcal{P}}$ and $\max_{\mathbf{V}}$ in Eq. (18), we obtain

$$\max_{\mathbf{V}} \min_{\mathbf{p}\in\mathcal{P}} f(\mathbf{V}, \mathbf{p}). \quad (19)$$

According to the minimax theorem [9], the optimal value of the objective function in Eq.(18) is an upper bound of that in Eq.(19).

The problem in Eq.(19) is a convex-concave optimization problem, and therefore its optimal solution is the saddle point of $f(\mathbf{V}, \mathbf{p})$ subject to the constraint in Eq.(17). Let $(\mathbf{V}^*, \mathbf{p}^*)$ be the optima to Eq.(19). For any feasible $\mathbf{p}$ and $\mathbf{V}$, we have

$$f(\mathbf{V}, \mathbf{p}^*) \leq f(\mathbf{V}^*, \mathbf{p}^*) \leq f(\mathbf{V}^*, \mathbf{p}). \quad (20)$$

To solve the problem in Eq.(19), one possible solution is to relax the indicator variable $p_i$ to $[0,1]$ and transform it into a multiple kernel learning problem [16]. It involves $d$ base kernels, where $d$ is the number of features. However, when the number of features is very huge, e.g., thousands, even the state of the art multiple kernel learning methods [16, 20] cannot handle it. Borrowing the idea used in [11, 19], we introduce an additional variable $\theta \in \mathbb{R}$, then the problem in Eq. (19) can be reformulated equivalently as follows

$$\max_{\mathbf{V}} \max_{\theta} -\theta$$
$$\text{s.t.} \theta \geq -f(\mathbf{V}, \mathbf{p}^t), \mathbf{p}^t \in \mathcal{P}. \quad (21)$$

Note that each $\mathbf{p}^t \in \mathcal{P}$ corresponds to one constraint, so the above optimization problem has $\binom{d}{m}$ constraints. The optimization problem in Eq.(21) is a Quadratically Constrained Linear Programming (QCLP) [3].

Taking the dual problem of the inner maximization problem in Eq(21), we obtain the following problem,

$$\max_{\mathbf{V}} \min_{\lambda_t \in \Lambda} \sum_{t=1}^{|\mathcal{P}|} \lambda_t f(\mathbf{V}, \mathbf{p}^t)$$
$$= \min_{\lambda_t \in \Lambda} \max_{\mathbf{V}} \sum_{t=1}^{|\mathcal{P}|} \lambda_t f(\mathbf{V}, \mathbf{p}^t), \quad (22)$$

where $\Lambda = \{\lambda_t | \sum_t \lambda_t = 1, \lambda_t \geq 0\}$. The equality holds due to the fact that the objective function is concave in $\mathbf{V}$ and convex in $\boldsymbol{\lambda}$.

### 3.4 Alternating Minimization

Eq.(22) can be seen as a multiple kernel learning problem [16]. Following the technique used in the state of the art multiple kernel learning [16], we optimize Eq. (22) in an alternative way. In particular, we alternatively solve $\mathbf{V}$ given the kernel weights $\boldsymbol{\lambda}$, and update the kernel weights $\boldsymbol{\lambda}$ by fixing $\mathbf{V}$. Let $g(\boldsymbol{\lambda}, \mathbf{V}) = \sum_{t=1}^{|\mathcal{P}|} \lambda_t f(\mathbf{V}, \mathbf{p}^t)$, we denote the gradient of $g(\boldsymbol{\lambda}, \mathbf{V})$ with respect to $\lambda_t$ by $\nabla_{\lambda_t} g(\boldsymbol{\lambda}, \mathbf{V})$, which is calculated as

$$\nabla_{\lambda_t} g(\boldsymbol{\lambda}, \mathbf{V}) = -\frac{1}{2\gamma}\text{tr}(\mathbf{V}^T \sum_{j=1}^{d} p_j^t \mathbf{K}_j \mathbf{V}). \quad (23)$$

Then we use projected gradient descent to update $\lambda_t$. The gradient of $g(\boldsymbol{\lambda}, \mathbf{V})$ with respect to $\mathbf{V}$ is

$$\nabla_{\mathbf{V}} g(\boldsymbol{\lambda}, \mathbf{V}) = \mathbf{H} - (\frac{1}{\gamma}\sum_{t=1}^{|\mathcal{P}|}\sum_{j=1}^{d} p_j^t \mathbf{K}_j + \mathbf{I})\mathbf{V}. \quad (24)$$

So $\mathbf{V}$ has a closed-form solution as

$$\mathbf{V} = (\frac{1}{\gamma}\sum_{t=1}^{|\mathcal{P}|}\lambda_t \sum_{j=1}^{d} p_j^t \mathbf{K}_j + \mathbf{I})^{-1}\mathbf{H}. \quad (25)$$

which can be solved as a linear system problem. One can also solve $\mathbf{W}$ in the primal and then compute $\mathbf{V}$ based on Eq.(10). The advantage is that the primal problem is a multivariate ridge regression, which can be solved very efficiently via iterative conjugate gradient type algorithms such as LSQR [13].

### 3.5 Cutting Plane Acceleration

Given $\mathcal{P}$, the above multiple kernel learning problem has optimization variables $(\mathbf{V}, \boldsymbol{\lambda})$ with $\binom{d}{m}$ constraints, which is impractical to solve. Fortunately, cutting plane technique [8] enables us to deal with this problem, which keeps a polynomial sized subset $\Omega$ of working constraints and computes the optimal solution to Eq. (22) subject to the constraints in $\Omega$. In detail, the algorithm adds the most violated constraint in Eq. (21) into $\Omega$ in each iteration. In this way, a successively strengthening approximation of the original problem is solved. And the algorithm terminates when no constraint in Eq. (21) is violated.

The remaining thing is how to find the most violated constraint in each iteration. Since the feasibility of a constraint is measured by the corresponding value of $\theta$, the most violated constraint is the one which owns the largest $\theta$. Hence, it could be calculated as follows

$$\arg\max_{\mathbf{p} \in \mathcal{P}} -f(\mathbf{V}, \mathbf{p})$$
$$= \arg\max_{\mathbf{p} \in \mathcal{P}} \mathrm{tr}(\mathbf{V}^T \sum_{j=1}^{d} p_j \mathbf{K}_j \mathbf{V})$$
$$= \arg\max_{\mathbf{p} \in \mathcal{P}} \sum_{j=1}^{d} s_j p_j, \qquad (26)$$

where $s_j = \mathbf{x}^j \mathbf{V}\mathbf{V}^T (\mathbf{x}^j)^T$. Note that its optimal solution can be obtained by first sorting $s_j$ and then setting the first $m$ numbers corresponding to $d_j$ to 1 and the rests to 0.

We summarize the algorithm to solve the problem in Eq. (22) in Algorithm 1. Note that the final selected features are the union set of the features corresponding to each constraint $\mathbf{p}^t \in \Omega_T$.

### 3.6 Theoretical Analysis

The convergence property of Algorithm 1 is stated in the following theorem.

**Theorem 3.3.** *Let $(\mathbf{V}^*, \theta^*)$ be the global optimal solution of Eq. (21), $l_t = \max_{1 \le j \le t} \min_{\mathbf{V}} -f(\mathbf{V}, \mathbf{p}^j)$ and $u_t = \min_{1 \le j \le t} \max_{\mathbf{p} \in \mathcal{P}} -f(\mathbf{V}^j, \mathbf{p})$, then*

$$l_t \le \theta^* \le u_t. \qquad (27)$$

*With the number of iteration t increasing, the sequence $\{l_t\}$ is monotonically increasing and the sequence $\{u_t\}$ is monotonically decreasing.*

---

**Algorithm 1** Generalized Fisher Score for Feature Selection

**Input:** $C$ and $m$;
**Output:** $\mathbf{V}$ and $\Omega$;
Initialize $\mathbf{V} = \frac{1}{n} \mathbf{1}_n \mathbf{1}_c^T$ and $t = 1$;
Find the most violated constraint $\mathbf{p}^1$, and set $\Omega_1 = \{\mathbf{p}^1\}$;
**repeat**
  Initialize $\boldsymbol{\lambda} = \frac{1}{t}\mathbf{1}$;
  **repeat**
    Solve for $\mathbf{V}$ using Eq.(25) under the current $\boldsymbol{\lambda}$;
    Solve for $\boldsymbol{\lambda}$ using gradient descent as in Eq. (23);
  **until** converge
  Find the most violated constraint $\mathbf{p}^{t+1}$ and set $\Omega_{t+1} = \Omega_t \cup \mathbf{p}^{t+1}$;
  $t = t + 1$;
**until** converge

---

In each outer iteration of our algorithm, it needs to find the most violated $\mathbf{p}$ and solve a multiple kernel learning problem. Finding the most violated $\mathbf{p}$ can be obtained exactly by finding the $m$ largest ones from $d$ coefficients $s_j$, which takes only $O(m \log d)$ time. To solve the multiple kernel learning problem, in each inner iteration, we need to solve one multivariate ridge regression problem, which can be solved efficiently by LSQR [13] and scales linearly in the number of training samples $n$. Hence the time complexity of multiple kernel learning is proportional to the complexity of ridge regression. In summary, the total time complexity of the proposed method is $O(T(cns + s \log m))$, where $T$ is the number of iterations needed to converge, $s$ is the average number of nonzero features among all the training samples, $c$ is the number of classes. In our experiments, 10 outer-iterations usually leads to converge. Thus, the proposed method is computationally very efficient.

## 4 Experiments

In our experiments, we empirically evaluate the effectiveness of the proposed method. We compare the proposed method to the state-of-art feature selection methods: Fisher Score [15], Laplacian Score [6], Hilbert Schmidt Independence Criterion (HSIC) [18] and Trace Ratio criterion [12]. Note that Trace Ratio criterion can use either Fisher score or Laplacian score like criteria. So we use Trace Ratio (FS) and Trace Ratio (LS) to represent them respectively. After feature selection, 1-Nearest Neighbor classifier is used for classification. The implementations of Laplacian score and Trace Ratio criterion are downloaded from the authors' websites. For HSIC, we use linear kernel on both data and labels. The parameters of the compared methods are tuned according to their original papers.

## 4.1 UCI Data Sets

In the first part of our experiments, we use a subset of UCI machine learning benchmark data set [2], e.g., ionosphere, sonar, protein and soybean.

Table.1 summarizes the characteristics of the data sets used in our experiments. All datasets are standardized to be zero-mean and normalized by standard deviation for each dimension. This normalization is also applied for the data used in the rest of our experiments.

Table 1: Description of the UCI data sets

| datasets | #samples | #features | #classes |
|---|---|---|---|
| ionosphere | 351 | 34 | 2 |
| sonar | 208 | 60 | 2 |
| protein | 116 | 20 | 6 |
| soybean | 307 | 35 | 19 |

We randomly choose 50% of the data for training and the rest for testing. Since the training samples are randomly chosen, we repeat this process 20 times and calculate the average result. The number of selected features is set to be 50% of the dimensionality of the data. Note that the number of selected features in GFS is controlled indirectly by $m$. We need to gradually increase $m$ to reach the chosen number of selected features. The regularization parameter $\gamma$ in GFS is tuned by 5-fold cross validation on the training set by searching the grid $\{50, 100, 200, \ldots, 500\}$. This parameter tuning approach is used throughout our experiments.

The classification results of the feature selection methods are summarized in Table 2. We observe that the proposed generalized Fisher score outperforms the other feature selection methods consistently on all the data sets. The improvement arises from two aspect: (1) GFS is able to consider the combination of features; and (2) it can handle redundant features. We will analyze it in more detail in the next part.

In addition, it is very interesting to find that there is no significant difference between Fisher score (or Laplacian score) and the corresponding Trace Ratio criterion. This is consistent with the observation in [24]. It is not surprising because both Trace Ratio criterion and Ratio Trace criterion essentially optimize quite similar objective functions. As far as we know, there is no theoretical evidence which supports that one of these two criteria is superior to the other.

Furthermore, although the performance of Fisher score is not as good as the proposed method, it is comparable to and even much better than the other feature selection methods on 3 out of 4 data sets. This indicates that Fisher score is still among the state of the art methods. It also implies the superiority of Fisher criterion for feature selection over the other criteria.

## 4.2 Face Recognition

In the second part of our experiments, we evaluate the proposed method on the ORL face recognition data set[1]. It contains 10 images for each of the 40 human subjects, which were taken at different times, varying the lighting, facial expressions and facial details. The original images (with 256 gray levels) have size $92 \times 112$, which are resized to $32 \times 32$ for efficiency. For each person, we randomly choose 5 images for training and the rest for testing. We repeat this experiment 20 times and calculate the average result.

The face recognition results of the feature selection methods when the number of selected features is 100 are shown in Table 3.

Table 3: Recognition results on the ORL data set when the number of selected features is 100

| Methods | Acc |
|---|---|
| HSIC | 74.47±3.08 |
| Fisher Score | 86.92±2.76 |
| Laplacian Score | 77.10±2.88 |
| Trace Ratio(FS) | 86.78±3.65 |
| Trace Ratio(LS) | 77.03±2.93 |
| GFS | **88.78±2.82** |

As can be seen, generalized Fisher score outperforms the other feature selection methods. To take a closer look at the performance with respect to the number of selected features, we plot the recognition accuracy with respect to the number of selected features of all the feature selection methods on the ORL data set in Figure 1. Since the number of selected features for the GFS is controlled by $m$, we increase $m$ gradually from 1 with step size 1 and obtain a increasing number of selected features.

We can see that with only a very small number of features, generalized Fisher score can achieve significant better result than the other methods. It can be interpreted from two aspects: (1) GFS selects features simultaneously, which considers the discriminative combination of features. For example, suppose feature combination $ab$ has a very high score, while feature

---
[1]http://www.cl.cam.ac.uk/Research/DTG/attarchive: pub/data

Table 2: Classification results on the UCI data sets when 50% data are used for training and the number of selected features is set to be 50% of the dimensionality of the data.

| Methods | ionosphere | sonar | protein | soybean |
|---|---|---|---|---|
| HSIC | 87.97±2.15 | 81.70±3.61 | 59.39±8.71 | 86.54±4.23 |
| Fisher Score | 87.97±1.96 | 81.31±3.48 | 67.63±6.77 | 76.31±3.28 |
| Laplacian Score | 83.29±2.10 | 80.87±3.51 | 67.19±6.64 | 78.10±3.77 |
| Trace Ratio(FS) | 88.23±2.32 | 81.36±3.19 | 67.72±6.52 | 76.54±3.80 |
| Trace Ratio(LS) | 83.66±2.48 | 81.07±3.50 | 68.60±6.05 | 77.91±3.34 |
| GFS | **89.14±2.02** | **82.33±3.97** | **69.21±5.87** | **87.06±2.50** |

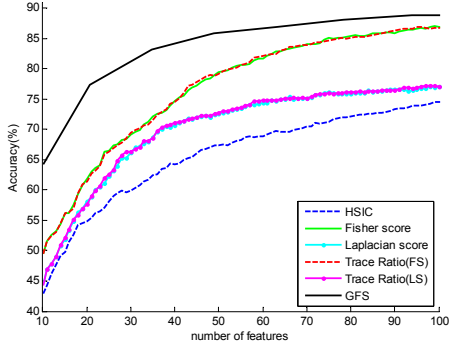

Figure 1: Recognition results of the feature selection methods with respect to the number of selected features on the ORL data set

$a$ and $b$ have relatively low scores respectively. Then GFS can select $ab$ at an early stage, while Fisher score as well as the other feature selection methods would select $a$ and $b$ respectively at a very late stage (i.e. until quite a lot of low-score features are selected); and (2) GFS is able to discard the redundant features, as a result, it can select as many as non-redundant features at an early stage.

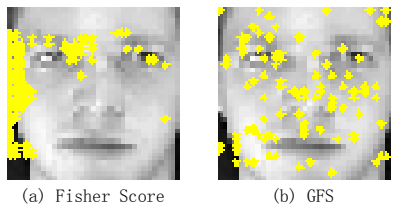

(a) Fisher Score   (b) GFS

Figure 2: The selected features by (a) Fisher score, and (b) GFS on the ORL data set

In order to give an intuitive picture, we display the first 100 features selected by Fisher score and the proposed GFS in Figure 2. It is shown that the distribution of selected features (pixels) by Fisher score is highly skewed. Most features distribute in the non-face region. It implies that the features selected by Fisher score are not discriminative. In contrast, the features selected by GFS distribute widely across the face region. Additionally, the selected features (pixels) are asymmetric and hence non-redundant. Since the face image is roughly axially symmetric, one pixel in a pair of axially symmetric pixels is redundant given the other one is selected. Furthermore, the selected pixels by GFS are mostly around the eyebrow, the corner of eyes, nose and mouth, which, in our experience, are more discriminative to distinguish face images of different people than those features selected by Fisher score. This is why GFS outperforms Fisher score.

### 4.3 Digit Recognition

In the third part of our experiments, we evaluate the proposed method on the USPS handwritten digit recognition data set [7]. A popular subset[2] containing 2007 $16 \times 16$ handwritten digit images is used in our experiments. We randomly choose 50% of the data for training and the rest for testing. This process is repeated 20 times.

Table 4 summarizes the digit recognition results of the feature selection methods when the number of selected features is 100, while Figure 3 depicts the classification accuracy with respect to the number of selected features of all the feature selection methods on the USPS data set.

Table 4: Recognition results on the USPS data set when the number of selected features is 100

| Methods | Acc |
|---|---|
| HSIC | 85.61±1.06 |
| Fisher Score | 91.85±0.82 |
| Laplacian Score | 83.90±1.05 |
| Trace Ratio(FS) | 91.83±0.81 |
| Trace Ratio(LS) | 83.33±1.51 |
| GFS | **92.69±1.16** |

Again, generalized Fisher score performs the best on this data set. Furthermore, GFS gets very good result even when the number of selected features is very small. For example, with only 10 features, GFS can achieve an accuracy of about 80%, while original Fisher score only gets roughly 50% accuracy.

---

[2]http://www-stat-class.stanford.edu/~tibs/ElemStatLearn/data.html

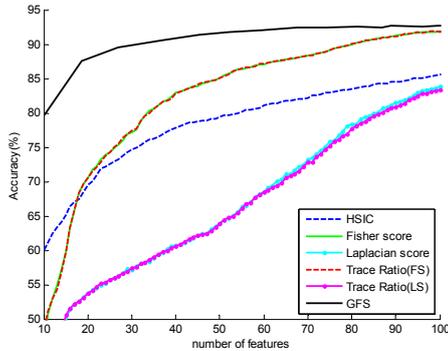

Figure 3: Recognition results of the feature selection methods with respect to the number of selected features on the USPS data set

This again strengthens the advantage of the proposed method.

## 5 Conclusion

In this paper, we presented a generalized Fisher score for feature selection. It finds a subset of features jointly, which maximize the lower bound of traditional Fisher score. The resulting feature selection problem is a mixed integer programming, which is reformulated as a quadratically constrained linear programming (QCLP). It can be solved by cutting plane algorithm, in each iteration of which a multiple kernel learning problem is solved by multivariate ridge regression and projected gradient descent. Experiments on benchmark data sets indicate that the proposed method outperforms many state of the art feature selection methods.

### Acknowledgments

The work was supported in part by NSF IIS-09-05215, U.S. Air Force Office of Scientific Research MURI award FA9550-08-1-0265, and the U.S. Army Research Laboratory under Cooperative Agreement Number W911NF-09-2-0053 (NS-CTA). We thank the anonymous reviewers for their helpful comments.